\documentclass[a4paper,fleqn,final, twocolumn]{cas-dc}

\usepackage[authoryear]{natbib}
\usepackage{placeins}
\usepackage{balance}

\def\tsc#1{\csdef{#1}{\textsc{\lowercase{#1}}\xspace}}
\tsc{WGM}
\tsc{QE}
\tsc{EP}
\tsc{PMS}
\tsc{BEC}
\tsc{DE}

\begin{document}
\let\WriteBookmarks\relax
\def\floatpagepagefraction{1}
\def\textpagefraction{.001}


\shorttitle{Vision-Language Model for Crop-Weed Segmentation }
\title[mode=title]{Vision-Language Semantic Grounding for Multi-Domain Crop-Weed Segmentation}

\shortauthors{Hossain et al.}

\author[a]{Nazia Hossain}

\author[a]{Xintong Jiang}

\author[a]{Yu Tian}

\author[b]{Philippe Seguin}

\author[a]{O. Grant Clark}

\author[a]{Shangpeng Sun}[orcid=0000-0001-7095-8626]
\ead{shangpeng.sun@mcgill.ca}


\affiliation[a]{organization={Department of Bioresource Engineering, McGill University},
  city={Sainte-Anne-de-Bellevue},
  state={QC},
  country={Canada}}

\affiliation[b]{organization={Department of Plant Science, McGill University},
  city={Sainte-Anne-de-Bellevue},
  state={QC},
  country={Canada}}

\begin{abstract}
Fine-grained crop-weed segmentation is essential for enabling targeted herbicide application in precision agriculture. However, existing deep learning models struggle to generalize across heterogeneous agricultural environments due to reliance on dataset-specific visual features. We propose Vision-Language Weed Segmentation (VL-WS), a novel framework that addresses this limitation by grounding pixel-level segmentation in semantically aligned, domain-invariant representations. Our architecture employs a dual-encoder design, where frozen Contrastive Language-Image Pretraining (CLIP) embeddings and task-specific spatial features are fused and modulated via Feature-wise Linear Modulation (FiLM) layers conditioned on natural language captions. This design enables image level textual descriptions to guide channel-wise feature refinement while preserving fine-grained spatial localization. Unlike prior works restricted to training and evaluation on single-source datasets, VL-WS is trained on a unified corpus that includes close-range ground imagery (robotic platforms) and high-altitude unmanned aerial vehicle (UAV) imagery, covering diverse crop types, weed species, growth stages, and sensing conditions. Experimental results across four benchmark datasets demonstrate the effectiveness of our framework, with VL-WS achieving a mean Dice score of 91.64\% and outperforming the strongest Convolutional Neural Network (CNN) baseline by 4.98\%. The largest gains occur on the most challenging weed class, where VL-WS attains 80.45\% Dice score compared to 65.03\% for the best baseline, representing a 15.42\% improvement. VL-WS further maintains stable weed segmentation performance under limited target-domain supervision, indicating improved generalization and data efficiency. These findings highlight the potential of vision-language alignment to enable scalable, label-efficient segmentation models deployable across diverse real-world agricultural domains.
\end{abstract}




\maketitle


\section{Introduction}
The objective of precision weed control is to minimize the impact of weed competition on the crop while reducing the amount of herbicide use through site-specific management. Central to this approach is accurate weed detection and localization, which enables targeted herbicide application that reduces chemical usage compared to broadcast spraying. Conventional blanket herbicide treatments not only increase costs but also accelerate environmental degradation, raise herbicide resistance, and deteriorate soil health \citep{jiang2020cnn, you2020dnn}. Pixel-level crop-weed segmentation offers the spatial accuracy required for localized spraying systems, making weed management economically viable and environmentally sustainable.
 
Modern crop-weed segmentation approaches predominantly rely on deep learning (DL) architectures that facilitate hierarchical representation learning and dense prediction in visually complex agricultural scenes. Encoder-decoder architectures such as U-Net and DeepLab variants have been widely applied to row-crop segmentation, demonstrating strong performance under complex field conditions with diverse weed species. Attention enhanced U-Net models have been used for sugar beet and sunflower fields, improving boundary delineation in cluttered scenes with overlapping vegetation \citep{li2025improved}. For maize, DeepLabV3+ has enabled real-time weed segmentation, demonstrating effectiveness under high inter-row variability and dense weed populations \citep{guo2025efficient}. With increasing use of unmanned aerial vehicle (UAV) imagery, convolutional neural network (CNN) and hybrid CNN-transformer architectures have been implemented for rice and soybean fields, where large-scale spatial context and variable lighting conditions pose challenges \citep{guo2025research,xu2023instance}. Transformer-based models and ConvNeXt-based backbones have shown improved robustness under changing weather, illumination, and weed density, while CycleGAN-based domain adaptation has been used to address seasonal and environmental shifts \citep{xu2025enhancing}. To reduce annotation costs, semi-supervised approaches have been applied to UAV imagery, achieving competitive performance using limited labeled data \citep{guo2025efficient}. Cross-domain transfer learning between ground-based and aerial platforms has further improved generalization across sensing modalities in the context of weed detection \citep{gao2024cross}.

Despite recent advances, a critical limitation restricts the real-world deployment of existing crop-weed segmentation models. Most models are trained and evaluated on a single dataset collected under specific field conditions, crop types, and sensing platforms. When applied to new agricultural environments where crop species, weed composition, growth stages, soil appearance, and imaging modalities vary substantially, these models often fail to generalize \citep{asad2024improved}. This limited generalization stems from a reliance on dataset-specific low-level visual cues related to plant morphology, such as texture, shape, and appearance patterns, rather than on higher-level semantic concepts of crops and weeds and, therefore, fails to transfer across agricultural domains. Creating a single dataset that captures the full diversity of real-world agricultural conditions is not practical. Comprehensive data collection would need to span entire growing seasons to capture phenological variation over time, while pixel-level annotation would become prohibitively expensive due to overlapping vegetation, micro-scale morphology, and ambiguous crop-weed boundaries. A more realistic solution is to leverage existing public datasets alongside limited domain-specific data, exposing the model to a broader range of crops, weed species, growth stages, and sensing conditions. Such multi-dataset training should enable the development of a single model that generalizes across diverse agricultural settings. However, naive aggregation of datasets often degrades performance due to semantic inconsistencies \citep{lambert2020mseg}. Shared labels such as ``weed'' group a broad range of morphologically distinct species under the same class, creating conflicting supervision signals that confuse models.

Recent advances in vision-language models show that integrating natural language with visual representations improves model robustness and generalization in complex visual tasks. In agricultural settings, vision-based models often struggle to generalize due to spatiotemporal variability, phenological changes, and environmental heterogeneity, which cause large appearance shifts across datasets \citep{wu2025fsvlm}. Language provides semantic priors that capture high-level concepts, such as spatial organization, growth stages, and contextual relationships that remain more consistent across environments than low-level visual cues. Prior work in agricultural vision-language studies show that language-guided models can effectively adapt to specialized tasks, such as plant stress phenotyping, disease diagnosis, and weed identification, even under limited supervision by facilitating the transfer of semantic knowledge across diverse agricultural conditions \citep{arshad2025leveraging}. Similarly, image-text paired agricultural datasets have shown that structured, domain-aware textual descriptions improve fine-grained classification by helping models distinguish between visually similar classes \citep{yu2025vl}. These findings motivate the exploration of semantic grounding in vision-language models to mitigate generalization challenges in crop-weed segmentation.

In this work, we explore vision-language semantic grounding to enable robust crop-weed segmentation across multiple datasets. We demonstrate that pretrained vision-language representations, organized around natural-language-aligned semantic concepts, provide a stable feature space that maintains semantic consistency across visually heterogeneous agricultural datasets. By grounding visual features in such language-aligned representations, the proposed approach reduces reliance on dataset-specific appearance patterns and mitigates negative transfer caused by semantic label heterogeneity. At the same time, a trainable segmentation-specific visual encoder and spatial decoder capture fine-grained spatial structure, enabling accurate delineation of crop-weed boundaries. In summary, we make the following contributions.

\begin{enumerate}
    \item We identify the limitations of standard CNN-based models in multi-dataset crop-weed segmentation and empirically show how inconsistent semantic labels degrade performance.
    \item We propose Vision-Language Weed Segmentation (VL-WS), a novel framework that integrates frozen Contrastive Language-Image Pretraining (CLIP) representations with a trainable spatial encoder, where Feature-wise Linear Modulation (FiLM)-based caption conditioning modulates fused features to achieve semantic stability across heterogeneous datasets varying in crop and weed species as well as ground sampling distance (GSD), while preserving precise boundary delineation.
    \item We validate the proposed approach on four diverse agricultural datasets, demonstrating improved performance in both cross-dataset generalization and data efficiency.
\end{enumerate}

\section{Related work}
\subsection{Challenges of Multi-Dataset Training with Shared Labels}
Recent research has explored multi-dataset training as a means of building more general and robust visual perception systems. \citet{zhou2022simple} demonstrates that labels with identical linguistic names may correspond to visually distinct concepts across datasets, and that naive dataset merging often leads to suboptimal performance. Their analysis further shows that effective multi-dataset learning requires dataset-aware training strategies to mitigate semantic conflicts and avoid performance degradation. Consistent with these findings, we observe a similar phenomenon in our crop-weed semantic segmentation experiments. Training standard CNNs jointly on multiple datasets degrades performance compared to dataset-specific training. This degradation persists despite the increased volume of training data, indicating that multi-dataset supervision introduces conflicting signals rather than providing complementary information. This behavior can be attributed to label-level semantic heterogeneity and negative transfer. As formalized by \citet{wang2019characterizing}, negative transfer arises when the divergence between source and target joint distributions is substantial and when learning algorithms fail to suppress incompatible source information. In our setting, the shared weed label aggregates approximately 12-14 distinct weed species across datasets, each exhibiting different morphology, texture, and growth characteristics. As shown by \citet{zhou2022simple} merging visually distinct concepts under a single semantic label causes label collisions that increase intra-class variability and destabilize feature learning. This semantic mismatch induces a conditional distribution shift between image features and labels, weakening discriminative representations and leading to consistent performance degradation in multi-dataset training.
\subsection{CLIP for Semantic Robustness in Multi-Dataset Learning}
\begin{figure}
    \centering
    \includegraphics[width=\columnwidth]{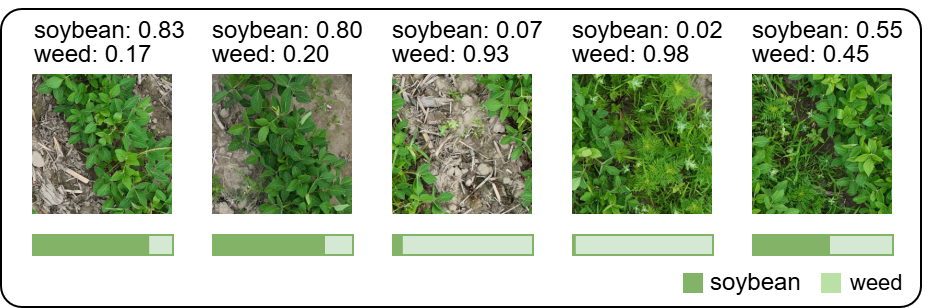}
    \caption{Image-text cosine similarity scores produced by a frozen CLIP model for soybean field images with different levels of weed presence. For crop-dominant scenes, image embeddings show higher similarity to the soybean prompt, while weed-dense scenes exhibit higher similarity to the weed prompt. This shift in similarity demonstrates that CLIP implicitly captures agronomic scene semantics without task-specific fine-tuning.}
    \label{fig:clip_similarity_weed}
\end{figure}

To address semantic conflicts and negative transfer in multi-dataset training, recent work has increasingly explored frozen pretrained encoder backbones \citep{bhattacharjee2023vision}. In particular, Contrastive Language-Image Pre-training (CLIP) has emerged as a prominent foundation model learning semantically structured visual representations through large-scale alignment of images and natural language descriptions \citep{radford2021learning}. Empirical studies show that language supervision enables CLIP to preserve meaningful intra-class feature variability, avoiding excessive feature collapse and improving robustness under heterogeneous data distributions \citep{wen2024makes}. By grounding visual representations in language rather than dataset-specific labels, CLIP induces a semantically structured feature space in which high-level concepts remain stable across domains and acquisition conditions \citep{bhalla2024interpreting}. This property is particularly advantageous in multi-dataset settings, where visually dissimilar instances may share a common label but differ substantially in appearance. As illustrated in Fig. \ref{fig:clip_similarity_weed}, pretrained CLIP image embeddings assign higher similarity to soybean prompts in crop-dominant scenes and to weed prompts in weed-dense images, demonstrating zero-shot semantic sensitivity to agricultural content.
\begin{figure}
    \centering
    \includegraphics[width=\columnwidth]{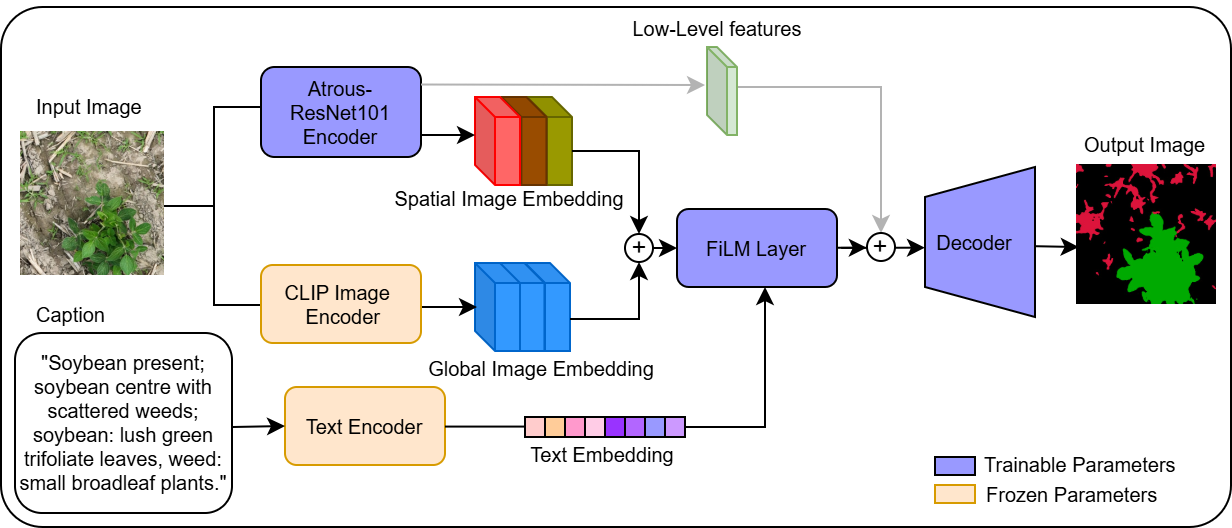}
    \caption{Overview of the proposed Vision-Language Weed Segmentation (VL-WS) framework. Dense spatial features extracted by a task-specific visual backbone are concatenated with global image embeddings from a pretrained CLIP encoder. The fused features are modulated by natural-language captions through Feature-wise Linear Modulation (FiLM), enabling text-conditioned channel adaptation for pixel-level crop-weed segmentation.}
    \label{fig:Architecture_overview.drawio}
\end{figure}
Unlike conventional CNN encoders trained under closed-vocabulary supervision, CLIP representations are open-vocabulary and less tied to dataset-specific visual statistics, reducing sensitivity to label noise and domain-specific shortcuts \citep{bourigault2025frevl}. Language supervision further encourages CLIP to organize features around semantic identity rather than low-level visual similarity, preserving meaningful intra-class variation under heterogeneous data distributions. As a result, frozen CLIP encoders mitigate negative transfer in multi-dataset training by enforcing domain-invariant, semantically grounded representations without requiring parameter adaptation. This semantic grounding is reflected in strong empirical performance, with frozen CLIP embeddings achieving 85-95\% of state-of-the-art accuracy on discriminative tasks without fine-tuning \citep{bourigault2025frevl}. However, CLIP embeddings remain inherently coarse and lack precise spatial localization, limiting their suitability for dense prediction tasks such as weed segmentation. 

\section{Materials and Methods}
To address the aforementioned limitations, we adopt a dual-encoder architecture (Fig. \ref{fig:Architecture_overview.drawio}) that combines frozen CLIP embeddings with a task-specific spatial encoder, which captures fine-grained texture, shape, and boundary information required for dense pixel-level segmentation across heterogeneous agricultural datasets. The high-level spatial features produced by the spatial encoder are fused with global CLIP image embeddings to form a multimodal representation. This fused feature map is then modulated by the caption embedding through FiLM \citep{perez2018film}, allowing semantic cues to selectively emphasize or suppress feature channels. The resulting FiLM-modulated features are fused with low-level spatial representations to recover fine boundaries and generate the final segmentation output. We validate our method using four heterogeneous agricultural datasets.

\subsection{Datasets and Annotations}
The study utilizes both a novel UAV soybean dataset and multiple publicly available agricultural benchmarks. This heterogeneous multi-dataset setting provides substantial variability in crop types and weed species, enabling robust performance analysis. In the following section, we provide detailed descriptions of data acquisition, annotation procedures, and characteristics for all datasets used in this study.
\subsubsection{UAV Soybean Dataset }
\begin{figure*}[t]
    \centering
    \includegraphics[width=0.6\textwidth]{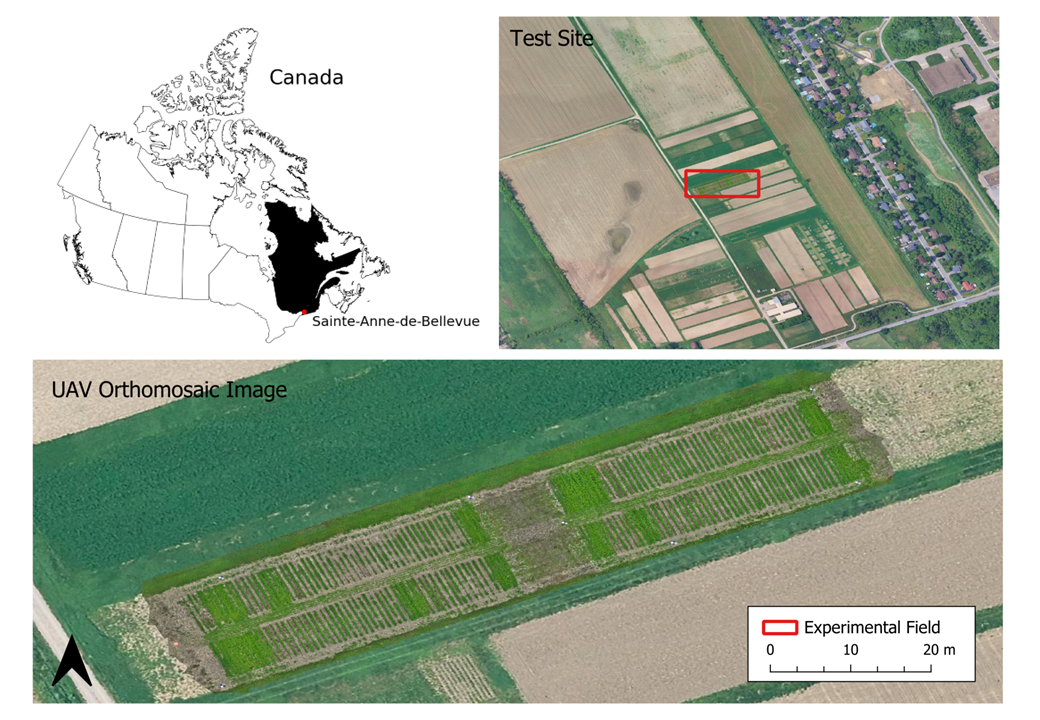}
    \caption{Study area and UAV orthomosaic of the experimental field. The figure illustrates the geographic context of the study site. The top-left panel shows the location of Sainte-Anne-de-Bellevue within the Province of Quebec, Canada, where the study area is located. The top-right panel presents an aerial view of the surrounding agricultural landscape, highlighting the specific experimental plot (outlined in red; 88~m $\times$ 18~m). The bottom panel displays the high-resolution UAV orthomosaic acquired using a DJI Mavic 3 Multispectral (M3M) at 5 m altitude, with a spatial scale bar provided for reference.}
    \label{fig:Uav}
\end{figure*}

We collected data from an actively managed soybean field at the Emile A. Lods Agronomy Research Centre of McGill University (Sainte-Anne-de-Bellevue, QC, Canada). We selected soybean as the target crop as it is a major field crop in Canada and soybean fields can often exhibit a high weed prevalence, presenting a representative scenario for crop-weed segmentation. An overview of the study area and the UAV-acquired orthomosaic of the experimental field are shown in Fig. \ref{fig:Uav}. We acquired UAV RGB imagery using a DJI Mavic 3 Multispectral (M3M) drone at an altitude of 5~m, covering an approximate area of \(88~\mathrm{m} \times 18~\mathrm{m}\) with a GSD  of 1.4~mm/pixel. The acquired images were processed to generate a georeferenced orthomosaic with uniform spatial resolution. We then partitioned the high-resolution orthomosaic into 7606 tiles of \(512 \times 512\) pixel with 25\% overlap to preserve the spatial continuity of crop and weed boundaries. Tiles containing more than 50\% non-informative pixels were discarded, and 293 images from the resulting set were manually annotated using LabelMe for semantic segmentation with three classes: background (soil), crop (soybean), and weed. For semantic segmentation, soybean plants were labeled as the \emph{crop} class, while all other vegetation was labeled as \emph{weed}. The resulting UAV Soybean-VL dataset builds upon the UAV Soybean dataset originally introduced in our prior work \citep{hossain2025end}. For this study, we apply a slightly revised data split by merging the original test set into the training set and reconstructing the validation set to emphasize more challenging weed density cases, while keeping the overall dataset composition unchanged. Under this protocol, the dataset comprises 262 training images and 30 validation images.

\subsubsection{Phenobench Dataset}
PhenoBench is a large, publicly available dataset and benchmark suite for plant perception under real agricultural field conditions \citep{10572312}. The dataset comprises 2872 high-resolution RGB image tiles (\(1024 \times 1024\) pixels) acquired using a UAV at a flight height of 21 m over sugar beet fields, resulting in a GSD of approximately 1~mm/pixel. Images were collected across multiple acquisition dates during a growing season, capturing substantial variability in plant growth stages and illumination conditions. Of the total dataset, 1407 images are designated for training and 772 images for validation and are publicly available, while the remaining data form a hidden test set used for standardized benchmarking. PhenoBench provides dense pixel-wise annotations for semantic segmentation, including soil, crop, weed, and partial-visibility classes, as well as instance segmentation of plants and leaves. Crop instances are temporally linked across acquisition dates, enabling longitudinal analysis of individual plant development. In addition to the labeled data, approximately 129000 unlabeled images are released to support research on self-supervised learning and domain adaptation in agricultural vision systems.

\subsubsection{GrowingSoy Dataset }
GrowingSoy is an instance-segmentation dataset for soybean and weed detection, consisting of 1000 manually annotated RGB images resized to \(640 \times 640\) pixels \citep{steinmetz2024seedling}. The images are extracted from 4K videos acquired in a dedicated soybean research field and span the full crop growth cycle, from early emergence to harvest. Data were collected at an experimental soybean plantation of the Universidade Federal de Santa Maria (Santa Maria, Brazil) using a 4K camera mounted on an all-terrain four-wheeled vehicle. The dataset includes soybean plants and common weed species, such as caruru and grassy weeds.

\subsubsection{ROSE Dataset}
\begin{figure*}[t]
    \centering
    \includegraphics[width=0.8\textwidth]{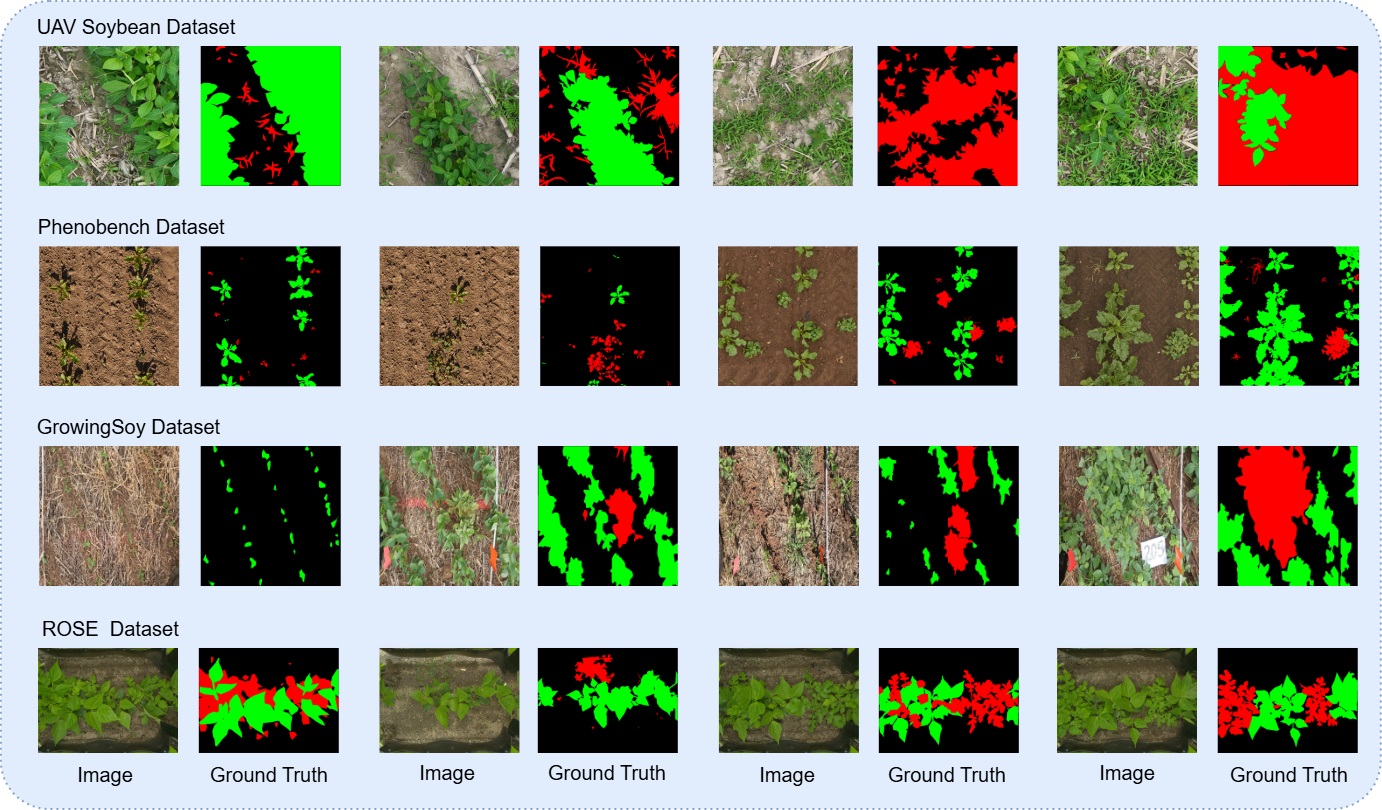}
    \caption{Representative RGB image tiles and corresponding ground-truth segmentation masks from the four weed segmentation datasets used in this study: UAV Soybean, PhenoBench, GrowingSoy, and ROSE. For each dataset, paired image-mask examples are shown, with RGB images on the left and manually annotated masks on the right. Mask colors denote background (black), crop (green), and weed (red). The datasets span diverse crop types, weed densities, and imaging conditions,, including aerial and ground-based imagery.}
    \label{fig:dataset}
\end{figure*}

The ROSE (RObotics and Sensors at the Service of Ecophyto) dataset is a multi-robot agricultural benchmark designed for weed segmentation, with a particular emphasis on robustness to environmental variability and domain shift \citep{avrin2020design}. It comprises images of maize (\textit{Zea mays}) and bean fields (\textit{Phaseolus vulgaris}) acquired by four different agricultural robots, introducing substantial variation in sensing modalities, camera configurations, and acquisition conditions. The dataset provides pixel-wise segmentation masks for crop and weed classes, with annotations covering multiple weed species that are merged into a single weed class for evaluation. ROSE is widely used to assess the generalization capability of segmentation models, including few-shot and domain-robust approaches, by training on data from multiple robots and evaluating on previously unseen robotic platforms. In our experiments, we used a subset of the dataset consisting of 250 images of bean crops and weeds. Representative image tiles and annotations from the UAV Soybean, PhenoBench, GrowingSoy, and ROSE datasets are shown in Fig.~\ref{fig:dataset}, highlighting the diversity in crop species, weed species, and acquisition conditions across datasets.

\subsection{Image Captions for Vision-Language Segmentation}
\begin{figure}
    \centering
    \includegraphics[width=\columnwidth]{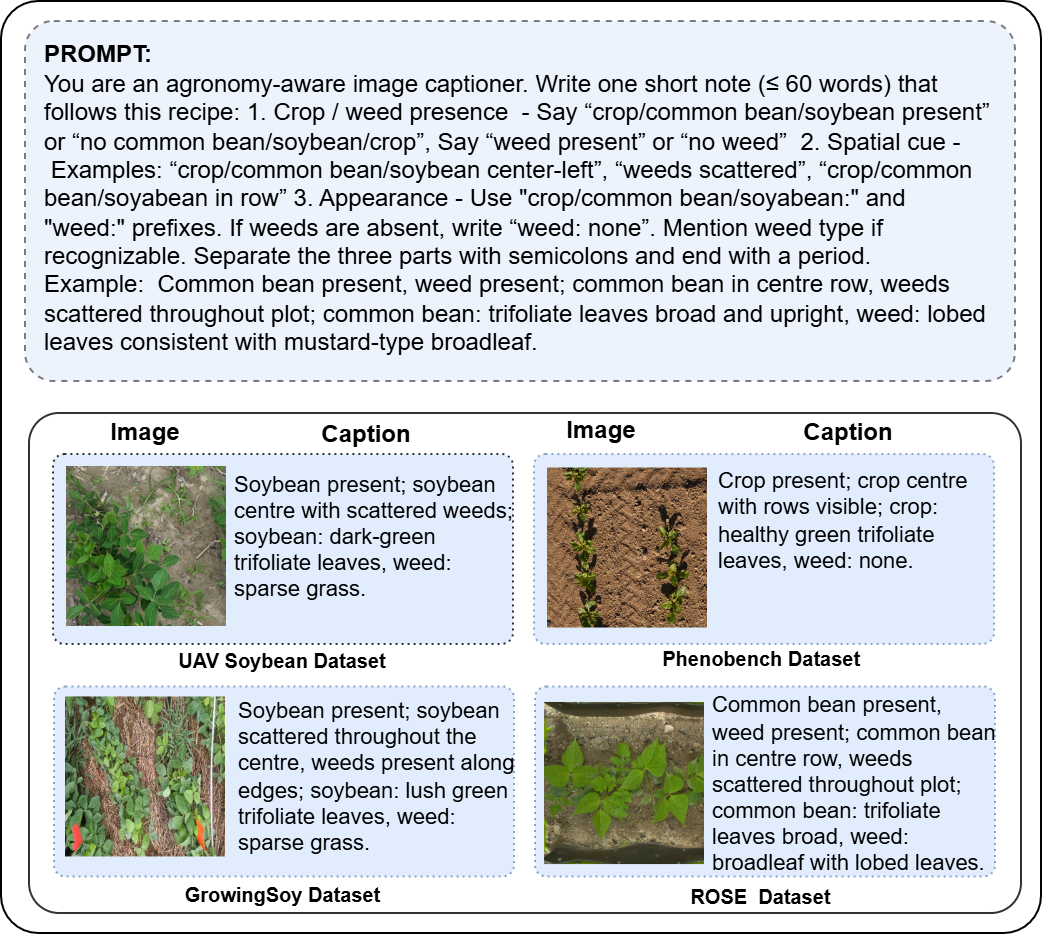}
    \caption{Example RGB image tiles paired with agronomy-aware natural language captions generated for each dataset (UAV Soybean, PhenoBench, GrowingSoy, and ROSE). Captions are produced using a standardized template and describe crop and weed presence, coarse spatial layout, and salient visual attributes. These image-caption pairs serve as multimodal inputs for training and evaluating the proposed vision-language segmentation framework. }
    \label{fig:dataset_caption.drawio}
\end{figure}

To enable vision-language learning, we pair each image tile from all datasets with an image-level, agronomy-aware caption. Captions are generated using a large language model (LLM) (GPT-4o-mini \citep{openai2024gpt4omini}) with a fixed prompt template that enforces a consistent structure across samples. Each caption describes the presence of crops and weeds, their coarse spatial arrangement, and salient visual characteristics observable in the image. This caption design is inspired by AgroGPT, which demonstrated that structured, domain-aware image descriptions with dataset specific attributes synthesized by LLMs for agricultural datasets can effectively inject expert knowledge into vision-language learning pipelines \citep{awais2025agrogpt}. Representative examples of the generated captions are shown in Fig.~\ref{fig:dataset_caption.drawio}. These image-caption pairs serve as multi-modal training inputs for the proposed VL-WS framework, ensuring semantic alignment between textual descriptions and visual content.

\subsection{Proposed Network Architecture}
\label{sec:proposed_arch}
\begin{figure*}[t]
    \centering
    \includegraphics[width=0.8\textwidth]{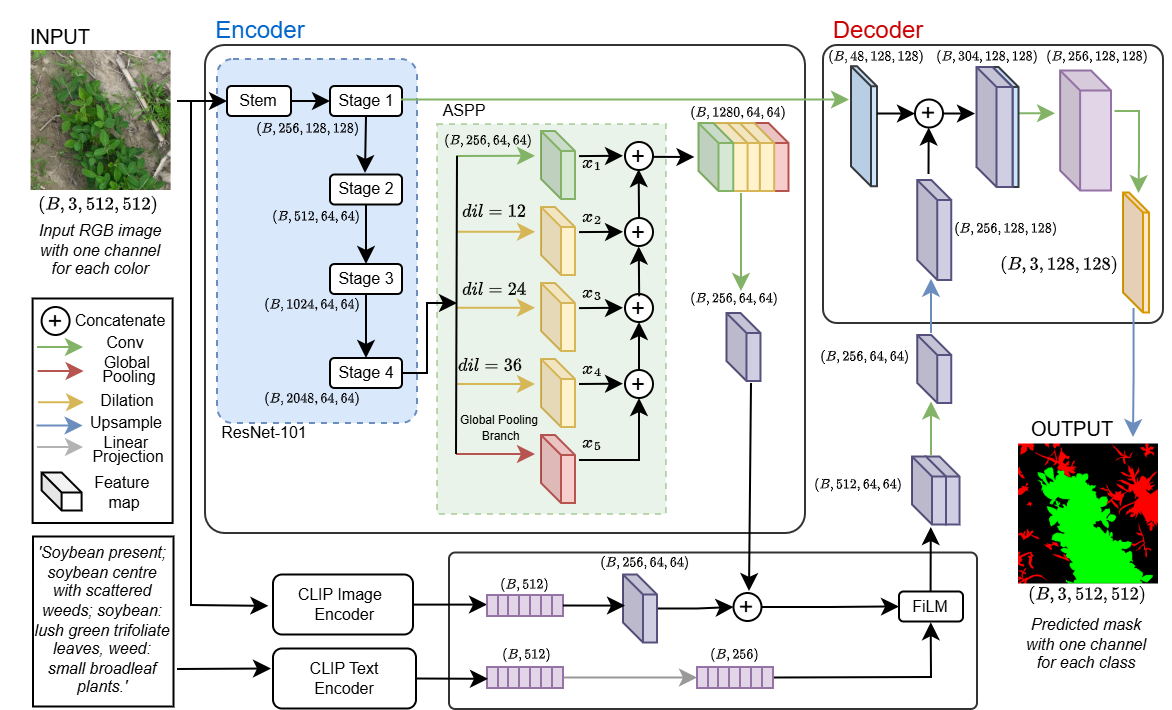}
    \caption{Detailed architecture of the proposed vision-language weed segmentation framework.}
    \label{fig:architecture_less_details}
\end{figure*}

\subsubsection{Overview}
The proposed framework, VL-WS, addresses the fundamental challenges of multi-dataset weed segmentation, including high intra-class variance due to morphologically diverse species and visual ambiguity between crops and weeds. Unlike conventional segmentation models that rely solely on visual features, VL-WS leverages frozen CLIP embeddings to ground pixel-level predictions in semantically meaningful concepts. As illustrated in Fig.~\ref{fig:architecture_less_details}, the architecture consists of two complementary visual encoding streams. A spatial backbone with atrous convolutions extracts multi-scale features while preserving fine-grained boundary details, whereas a frozen CLIP image encoder provides global semantic representations that remain stable across diverse agricultural domains. These visual features are subsequently fused and conditioned using agronomy-aware captions through FiLM. Text embeddings modulate feature channels to emphasize semantically relevant patterns and reduce sensitivity to domain-specific variation.

\subsubsection{Visual Encoder Backbone }
The spatial encoder in the VL-WS model is built on the DeepLabv3+ architecture. We adopt a ResNet-101 backbone with atrous convolutions configured for an output stride of 8 for preserving fine spatial detail while capturing rich semantic context. This higher-resolution representation is important for distinguishing morphologically similar crop and weed structures in agricultural images. In its standard form, ResNet-101 reduces spatial resolution by a factor of 32 through successive pooling and strided convolutions. Such aggressive downsampling eliminates fine details essential for boundary delineation in dense prediction tasks. To mitigate this limitation, DeepLabv3+ replaces strided convolutions in the last two residual blocks with atrous (dilated) convolutions, resulting in effective dilation rates of 2 and 4 in the final encoder stages. This design maintains a large receptive field while preserving feature maps at one-eighth of the input resolution.

The encoder outputs a hierarchical set of feature maps at different scales. A low-level feature map
$F_L \in \mathbb{R}^{256 \times \frac{H}{4}\times\frac{W}{4}}$,
extracted from early convolutional layers, retains fine spatial detail essential for boundary refinement. In contrast, the deepest feature map
$F_D \in \mathbb{R}^{2048 \times \frac{H}{8}\times\frac{W}{8}}$,
produced after stage 4, captures high-level semantic context. This high-level representation is then passed through the Atrous Spatial Pyramid Pooling (ASPP) module, which aggregates multi-scale context through parallel atrous convolutions with dilation rates of 12, 24, and 36. The ASPP module also includes a $1\times1$ convolution branch and an image-level pooling branch that performs global average pooling followed by a $1\times1$ convolution and bilinear upsampling. The outputs from all ASPP branches are concatenated and projected through a final $1\times1$ convolution, yielding $F_{\mathrm{ASPP}} \in \mathbb{R}^{256 \times \frac{H}{8}\times\frac{W}{8}}$.

\subsubsection{Language and Image Embeddings via CLIP}
CLIP is a large-scale vision-language model trained on paired image-text data to learn a shared embedding space that aligns visual content with linguistic semantics. This joint training provides CLIP with strong semantic priors that are transferable across domains and robust to appearance variation. We leverage these properties to inject high-level semantic context into pixel-level segmentation using CLIP-derived global features and text embeddings. 

\paragraph{Global Image Encoding.}
The input image is resized to \(224 \times 224\) and passed through the CLIP image encoder to obtain a 512-dimensional global embedding. This embedding is linearly projected to 256 dimensions, yielding \(E_{\mathrm{vis}} \in \mathbb{R}^{256}\), which captures scene-level semantics that are less sensitive to local texture variation or sensor noise. To preserve the pretrained vision-language alignment, the CLIP image encoder is kept fully frozen during training, with only the projection layer being optimized.

\paragraph{Text Encoding.}
In parallel with global image encoding, the corresponding natural language caption (e.g., ``Soybean center with scattered weeds'') is tokenized and processed by the CLIP text encoder to produce a 512-dimensional embedding. This embedding is then projected to 256 dimensions, producing \(E_{\mathrm{txt}} \in \mathbb{R}^{256}\). To adapt the representation to domain-specific agricultural terminology while retaining general language understanding, the final two transformer layers of the text encoder are fine-tuned during training.

\subsubsection{Multimodal Fusion and FiLM Modulation}
This module fuses dense spatial features with global visual embeddings and applies caption-conditioned modulation to guide segmentation. The ASPP feature map $F_{\mathrm{ASPP}} \in \mathbb{R}^{256 \times \frac{H}{8} \times \frac{W}{8}}$ is concatenated channel-wise with the CLIP image embedding $E_{\mathrm{vis}} \in \mathbb{R}^{256}$, which is spatially broadcast to match the ASPP resolution, yielding the fused representation $F_{\mathrm{fused}} \in \mathbb{R}^{512 \times \frac{H}{8} \times \frac{W}{8}}$. Feature-wise Linear Modulation (FiLM) is then applied to dynamically condition the fused features on the input caption. The text embedding $E_{\mathrm{txt}} \in \mathbb{R}^{256}$, obtained by linearly projecting the CLIP text encoder output, is used to generate element-wise scaling $\gamma(E_{\mathrm{txt}})$ and shifting $\beta(E_{\mathrm{txt}})$ parameters for the FiLM operation, which modulate the fused features as
\begin{equation}
\tilde{F} = \gamma(E_{\mathrm{txt}}) \odot F_{\mathrm{fused}} + \beta(E_{\mathrm{txt}}).
\end{equation}

This caption-conditioned modulation selectively emphasizes feature channels aligned with the semantic content of the caption and guides the decoder toward semantically consistent segmentation across heterogeneous datasets.

\subsubsection{Decoder}
The modulated features are refined using a DeepLabv3+ style decoder to recover precise object boundaries. First, $\tilde{F}$ is bilinearly upsampled by a factor of 2 and concatenated with low-level features from the backbone, which are compressed to 48 channels via a $1\times1$ convolution. This fusion produces a 304-channel representation that is processed by two successive $3 \times 3$ convolutions, reducing the channel dimensionality to 256. A final $1 \times 1$ convolution generates logits for the three semantic classes: background, crop, and weed.

\subsection{Loss Function}

VL-WS is trained using a composite loss that combines a weighted segmentation loss with a vision-language contrastive loss, enabling accurate pixel-level predictions while enforcing alignment between spatial features and textual semantics.

\subsubsection{Segmentation Loss}

To supervise per-pixel classification, we adopt a hybrid Dice + Cross-Entropy loss to balance region-level overlap and pixel-wise correctness. The total segmentation loss is defined as:

\begin{equation}
\mathcal{L}_{\text{seg}} = 
\lambda_{\text{Dice}} \cdot \mathcal{L}_{\text{Dice}} +
\lambda_{\text{CE}} \cdot \mathcal{L}_{\text{CE}},
\end{equation}
where we set $\lambda_{\text{Dice}} = 0.6$ and $\lambda_{\text{CE}} = 0.4$.  
Both loss components operate over class probabilities obtained via softmax.

\paragraph{Dice Loss.}
Dice Loss encourages region-level agreement, especially useful in imbalanced datasets (e.g., weed vs. background):

\begin{equation}
\mathcal{L}_{\text{Dice}}
=
1 - \frac{1}{C}
\sum_{c=1}^{C}
\frac{
2 \sum_{i=1}^{N} p_{ic} y_{ic}
}{
\sum_{i=1}^{N} p_{ic}
+
\sum_{i=1}^{N} y_{ic}
+ \varepsilon
}.
\end{equation}

Here, \(C\) denotes the number of semantic classes, and \(N\) is the total number of pixels. For each pixel \(i\) and class \(c\), \(p_{ic}\) represents the predicted softmax probability, while \(y_{ic} \in \{0,1\}\) is the corresponding one-hot ground-truth label. A small constant \(\varepsilon\) is added for numerical stability.

\paragraph{Cross-Entropy Loss.}
Cross-Entropy Loss penalizes pixel-wise misclassifications by comparing the predicted class masks against the one-hot encoded ground-truth labels. For a multi-class segmentation task with $C$ classes and $N$ pixels, the weighted categorical cross-entropy loss is defined as:

\begin{equation}
\mathcal{L}_{\text{CE}}(y, p)
=
- \frac{1}{N}
\sum_{i=1}^{N}
\sum_{c=1}^{C}
w_c \, y_{i,c} \, \log(p_{i,c}),
\end{equation}
where $y_{i,c}$ denotes the one-hot ground-truth label for pixel $i$ and class $c$, and $p_{i,c}$ is the predicted probability for that class obtained from the softmax output. The class-specific weight $w_c$ is used to mitigate class imbalance by assigning a larger penalty to minority classes. 

\subsubsection{Vision-Language Contrastive Loss}

To reinforce semantic alignment between textual labels and global image features, we incorporate a symmetric InfoNCE loss following \citet{radford2021learning} and \citet{zhang2022contrastive}.  
Given a batch of $N$ paired image-text embeddings $\{(v_i, t_i)\}_{i=1}^N$, where $v_i$ and $t_i$ denote the visual and textual representations, we first normalize both embeddings:

\begin{align}
\hat{v}_i &= \mathrm{normalize}\big(\mathrm{mean}(f_i^{\text{img}})\big), & \text{(visual embedding)} \\
\hat{t}_i &= \mathrm{normalize}(f_i^{\text{text}}), & \text{(text embedding)}
\end{align}

The image-to-text and text-to-image contrastive objectives are defined as:
\begin{align}
\ell_i^{(v \rightarrow t)} &= 
- \log 
\frac{
\exp(\langle \hat{v}_i, \hat{t}_i \rangle / \tau)
}{
\sum_{k=1}^{N} \exp(\langle \hat{v}_i, \hat{t}_k \rangle / \tau)
}, \\
\ell_i^{(t \rightarrow v)} &= 
- \log 
\frac{
\exp(\langle \hat{t}_i, \hat{v}_i \rangle / \tau)
}{
\sum_{k=1}^{N} \exp(\langle \hat{t}_i, \hat{v}_k \rangle / \tau)
},
\end{align}
where $\langle \hat{v}, \hat{t} \rangle = \hat{v}^\top \hat{t}$ denotes cosine similarity and $\tau$ is a temperature hyper-parameter set to 0.07.

The final symmetric contrastive loss is averaged across both modalities:
\begin{equation}
\mathcal{L}_{\text{VL}} =
\frac{1}{2N}
\sum_{i=1}^{N}
\big(
\ell_i^{(v \rightarrow t)} + \ell_i^{(t \rightarrow v)}
\big).
\end{equation}

This bidirectional InfoNCE objective encourages each image embedding to be most similar to its corresponding textual description while remaining dissimilar to other texts within the batch, promoting semantically aligned feature representations.

\subsubsection{Total Loss}

The final loss function combines segmentation and vision-language components:

\begin{equation}
\mathcal{L}_{\text{total}} =
\mathcal{L}_{\text{seg}} +
\lambda_{\text{VL}} \cdot \mathcal{L}_{\text{VL}},
\qquad
\lambda_{\text{VL}} = 0.02
\end{equation}

The contrastive term acts as an auxiliary supervision signal that enhances the semantic expressiveness of the visual encoder and improves model generalization across heterogeneous domains.

\subsection{Implementation Details}

\paragraph{Evaluation Metric.}
Segmentation performance is primarily evaluated using the Dice coefficient, which measures the spatial overlap between predicted and ground-truth regions. For a given class $c$, the Dice score is defined as
\begin{equation}
\mathrm{Dice}_c = \frac{2 \lvert P_c \cap G_c \rvert}{\lvert P_c \rvert + \lvert G_c \rvert},
\end{equation}
where $P_c$ and $G_c$ denote the sets of pixels predicted as and labeled as class $c$, respectively. Dice is computed globally per class across the validation set and then averaged over classes. This metric is particularly well suited for crop-weed segmentation due to its robustness to class imbalance and its emphasis on accurate region-level overlap.

\paragraph{Training details.} 
All experiments were conducted on a workstation equipped with an NVIDIA RTX~3080 GPU (10 GB). Input images are resized to $512 \times 512$ pixels, and the model is implemented in PyTorch using PyTorch Lightning. We employ the AdamW optimizer with an initial learning rate of $3 \times 10^{-5}$ for the visual encoder and $3 \times 10^{-6}$ for the text encoder. Training is performed for 200 epochs using a cosine annealing learning rate schedule with a minimum learning rate of $1 \times 10^{-7}$, a batch size of 8 and early stopping with a patience of 30 epochs.

\paragraph{Baselines.}
We evaluate our approach against three widely used semantic segmentation baselines including DeepLabv3+ \citep{chen2018encoder}, U-Net \citep{ronneberger2015u}, and PSPNet \citep{zhao2017pyramid}. DeepLabv3+ is selected as the primary baseline due to its strong ability to capture multi-scale context, while U-Net and PSPNet provide representative encoder-decoder and pyramid pooling architectures. These baselines are chosen to represent strong unimodal CNN-based methods commonly used in agricultural semantic segmentation.

\section{Results and Discussion}

\subsection{Comparative Analysis with Standard Baselines}
All models were trained and evaluated in a multi-dataset setting comprising UAV Soybean, PhenoBench, GrowingSoy, and ROSE. This experimental setup reflects a realistic deployment scenario characterized by substantial domain variability in crop types, weed species, imaging conditions, and sensing platforms.

\begin{table}[h]
\centering
\caption{Aggregated segmentation performance (Dice \%) across all datasets.}
\label{tab:overall_singlecol}
\begingroup
\setlength{\tabcolsep}{3pt}
\begin{tabular}{lcccc}
\toprule
\textbf{Method} &
\textbf{\shortstack{Weed}} &
\textbf{\shortstack{Crop}} &
\textbf{\shortstack{Bg}} &
\textbf{\shortstack{Avg}} \\
\midrule
UNet                     & 61.45 & 95.73 & 99.56 & 85.58 \\
PSPNet                   & 63.47 & 93.04 & 99.08 & 85.19 \\
DeepLabv3+               & 65.03 & 95.48 & 99.47 & 86.66 \\
VL-WS & \textbf{80.45} & 95.23 & 99.24 & \textbf{91.64} \\
\bottomrule
\end{tabular}
\endgroup
\end{table}

\begin{table*}[t]
\centering
\caption{Dataset-wise segmentation performance (Dice \%) of the proposed model and baseline methods across four agricultural datasets.}
\label{tab:dataset_wise}
\setlength{\tabcolsep}{4pt}
\begin{tabular}{lcccccccccccc}
\toprule
\textbf{Method} &
\multicolumn{3}{c}{\textbf{UAV Soybean}} &
\multicolumn{3}{c}{\textbf{PhenoBench}} &
\multicolumn{3}{c}{\textbf{GrowingSoy}} &
\multicolumn{3}{c}{\textbf{ROSE}} \\
\cmidrule(lr){2-13}
& \textbf{Weed} & \textbf{Crop} & \textbf{Bg} &
  \textbf{Weed} & \textbf{Crop} & \textbf{Bg} &
  \textbf{Weed} & \textbf{Crop} & \textbf{Bg} &
  \textbf{Weed} & \textbf{Crop} & \textbf{Bg} \\
\midrule
\shortstack[l]{UNet} &
79.02 & 98.12 & 96.73 &
58.27 & 96.57 & 99.70 &
75.84 & 92.62 & 98.80 &
43.97 & 88.25 & 97.11 \\
PSPNet &
69.34 & 97.28 & 93.98 &
59.07 & 93.29 & 99.22 &
76.42 & 91.22 & 98.35 &
62.53 & 90.69 & 96.75 \\
Deeplabv3+ &
74.48 & 98.12 & 95.53 &
60.33 & 96.10 & 99.62 &
75.30 & 92.51 & 98.65 &
65.19 & 90.92 & 96.85 \\
{VL-WS} &
\textbf{80.40} & 98.21 & 96.47 &
\textbf{77.57} & 95.84 & 99.49 &
\textbf{86.09} & 92.98 & 98.66 &
\textbf{75.66} & 92.49 & 97.30 \\
\bottomrule
\end{tabular}
\end{table*}

\subsubsection{Aggregated Performance Analysis}
The aggregated quantitative results across the multi-domain test set are reported in Table~\ref{tab:overall_singlecol}. VL-WS achieves competitive performance relative to state-of-the-art segmentation models under multi-dataset training conditions, attaining a mean Dice score of $91.64\%$ across the weed, crop, and background classes. VL-WS outperforms the strongest unimodal baseline, DeepLabv3+, which achieves an average Dice score of $86.66\%$, yielding an absolute improvement of $4.98\%$. Other CNN-based architectures exhibit notably lower performance, with U-Net achieving $85.58\%$ and PSPNet achieving $85.19\%$ mean Dice scores. These results highlight the limitations of purely visual encoders in multi-domain agricultural settings, where domain shift and high intra-class variability present significant challenges.

CNN-based segmentation models, such as DeepLabv3+ and PSPNet, incorporate architectural mechanisms for contextual reasoning, ASPP and pyramid pooling modules, respectively. However, these purely visual mechanisms prove to be insufficient for robust cross-domain generalization. Models that rely exclusively on spatial features and low-level visual patterns struggle to distinguish fine-grained similarities between morphologically diverse weed species and young crop plants. This limitation is most evident in the performance of the weed class.

\subsubsection{Class-Wise Performance Analysis}
A class-wise analysis provides deeper insight into the sources of performance improvement, revealing that the gains of VL-WS are concentrated in the most challenging semantic category.

\paragraph{Weed Class Performance.}
The weed class exhibits the largest performance disparity between VL-SM and baseline methods, reflecting the inherent difficulty of weed segmentation in precision agriculture. VL-WS achieves an average of weed Dice score of $80.45\%$, substantially outperforming DeepLabv3+ ($65.03\%$), PSPNet ($63.47\%$), and U-Net ($61.45\%$)  across the four datasets (Table~\ref{tab:overall_singlecol}). This corresponds to a $15.42\%$ improvement over DeepLabv3+. The difficulty of weed segmentation stems from multiple compounding factors. During early growth stages, crops and weeds exhibit highly similar spectral and morphological characteristics, limiting discriminative visual cues and increasing pixel-level ambiguity. Moreover, the weed class exhibits substantially higher intra-class variance than crop. Although the crop class typically corresponds to a single species per dataset (e.g. soybean in UAV Soybean or common bean in ROSE), the weed class aggregates multiple distinct species even within a single dataset. Across the combined dataset benchmark, weed class encompasses approximately 12-14 morphologically and spectrally distinct weed types. This high intra-class heterogeneity exacerbates the generalization challenge. The pronounced performance gap between the VL-WS and standard CNN baselines suggests that conventional architectures suffer from negative transfer when trained jointly on heterogeneous multi-domain datasets. These models overfit superficial domain-specific cues (e.g. texture or sensor noise) instead of learning semantically grounded, generalizable plant representations.

\paragraph{Crop Class Performance.}
All models achieve strong and comparable performance in the crop class. U-Net achieves a Dice score of $95.73\%$, DeepLabv3+ achieves $95.48\%$, VL-WS achieves $95.23\%$, and PSPNet achieves $93.04\%$. The marginal performance differences among the top-performing models indicate that crop segmentation has largely saturated under current evaluation settings, offering minimal scope for further improvement.

\paragraph{Background Class Performance.}
All models achieve near-ceiling performance on the background class, with Dice scores of $99.56\%$ (U-Net), $99.47\%$ (DeepLabv3+), $99.24\%$ (VL-SM), and $99.08\%$ (PSPNet). This suggests that background-vegetation discrimination has saturated under current segmentation architectures. Consequently, the substantial performance gains of the proposed VL-WS over baseline methods arise primarily from improved weed segmentation, rather than marginal improvements in crop or background classification.

\subsubsection{Dataset-Wise Weed Segmentation Performance}
Across all four benchmark datasets (UAV Soybean, PhenoBench, GrowingSoy, and ROSE), the VL-WS model consistently outperforms all baseline methods on weed segmentation, as summarized in Table~\ref{tab:dataset_wise}. On UAV Soybean, VL-WS achieves a weed Dice score of \(80.40\%\), exceeding DeepLabv3+, PSPNet, and U-Net by margins of \(5.92\%\), \(11.06\%\) \, and \(1.38\%\), respectively. The performance gap is more pronounced on PhenoBench, where VL-WS attains a weed Dice score of \(77.57\%\), substantially outperforming all comparison models. VL-WS achieves its highest weed Dice score on GrowingSoy (\(86.09\%\)), surpassing DeepLabv3+ by \(10.79\%\). On the ROSE dataset, VL-WS maintains robust performance with a weed Dice score of \(75.66\%\), outperforming all baselines. Moreover, VL-WS exhibits the smallest variance in weed segmentation performance across datasets (\(75.66\%\)-\(86.09\%\)) compared to DeepLabv3+, U-Net, and PSPNet, demonstrating superior cross-domain consistency.

\subsection{Domain Adaptation and Sample Efficiency}
To assess the data efficiency of the VL-WS model, we evaluate its performance under varying levels of target-domain supervision. For each target dataset, the model is trained using the full training sets of the three remaining datasets together with different fractions of labeled data from the target domain (10\%, 20\%, 50\%, and 100\%). The validation split of the target dataset is kept fixed throughout. This evaluation protocol reflects realistic deployment scenarios in which models pre-trained on auxiliary domains must adapt to new environments under limited annotation budgets. Table~\ref{tab:domain_adaptation} reports class-wise Dice scores across all four target datasets under different data availability regimes. VL-WS maintains strong performance even under minimal supervision. In particular, weed segmentation accuracy remains close to the full-data setting when trained with 50\% of the target-domain annotations and then degrades gradually as the amount of target-domain data is further reduced. Background and crop classes show minimal sensitivity to data availability, with Dice scores consistently exceeding 90\% across most settings. Overall, these results demonstrate the data efficiency and cross-domain generalization capability of the proposed framework, highlighting its practical applicability in agricultural scenarios where labeled data are limited.
\begin{table*}[t]
\centering
\caption{Domain adaptation performance under limited target-domain supervision. Class-wise Dice (\%) scores obtained by training on three full source datasets and varying fractions of labeled target-domain data.}
\label{tab:domain_adaptation}
\begin{tabular*}{\textwidth}{@{\extracolsep{\fill}}llccccc@{}}
\toprule
\multirow{2}{*}{\textbf{Target Dataset}} &
\multirow{2}{*}{\textbf{Source Datasets}} &
\multirow{2}{*}{\textbf{Class}} &
\multicolumn{4}{c}{\textbf{Target-Domain Training Data}} \\
& & & \textbf{10\%} & \textbf{20\%} & \textbf{50\%} & \textbf{100\%} \\
\midrule
\multirow{4}{*}{UAV Soybean}
 & \multirow{4}{*}{\shortstack{PhenoBench + GrowingSoy\\+ ROSE}}
 & Weed       & 68.63 & 74.63 & 79.95 & 80.40 \\
 & & Crop       & 96.92 & 97.31 & 98.06 & 98.21 \\
 & & Background & 94.80 & 95.95 & 96.37 & 96.47 \\
 & & Mean       & 86.78 & 89.29 & 91.46 & 91.69 \\
\midrule
\multirow{4}{*}{PhenoBench}
 & \multirow{4}{*}{\shortstack{UAV Soybean + GrowingSoy\\+ ROSE}}
 & Weed       & 69.94 & 73.66 & 75.71 & 77.57 \\
 & & Crop       & 94.18 & 95.26 & 95.59 & 95.84 \\
 & & Background & 99.30 & 99.43 & 99.46 & 99.49 \\
 & & Mean       & 87.80 & 89.45 & 90.25 & 90.96 \\
\midrule
\multirow{4}{*}{GrowingSoy}
 & \multirow{4}{*}{\shortstack{UAV Soybean + PhenoBench\\+ ROSE}}
 & Weed       & 78.21 & 81.20 & 84.37 & 86.09 \\
 & & Crop       & 90.99 & 91.74 & 92.14 & 92.98 \\
 & & Background & 98.40 & 98.49 & 98.53 & 98.66 \\
 & & Mean       & 89.20 & 90.47 & 91.68 & 92.57 \\
\midrule
\multirow{4}{*}{ROSE}
 & \multirow{4}{*}{\shortstack{UAV Soybean + PhenoBench\\+ GrowingSoy}}
 & Weed       & 61.89 & 70.78 & 72.42 & 75.66 \\
 & & Crop       & 87.15 & 91.54 & 92.14 & 92.49 \\
 & & Background & 95.99 & 97.09 & 97.27 & 97.30 \\
 & & Mean       & 81.67 & 86.47 & 87.27 & 88.48 \\
\bottomrule
\end{tabular*}
\end{table*}

\begin{figure*}[h]
    \centering
    \includegraphics[width=0.8\textwidth]{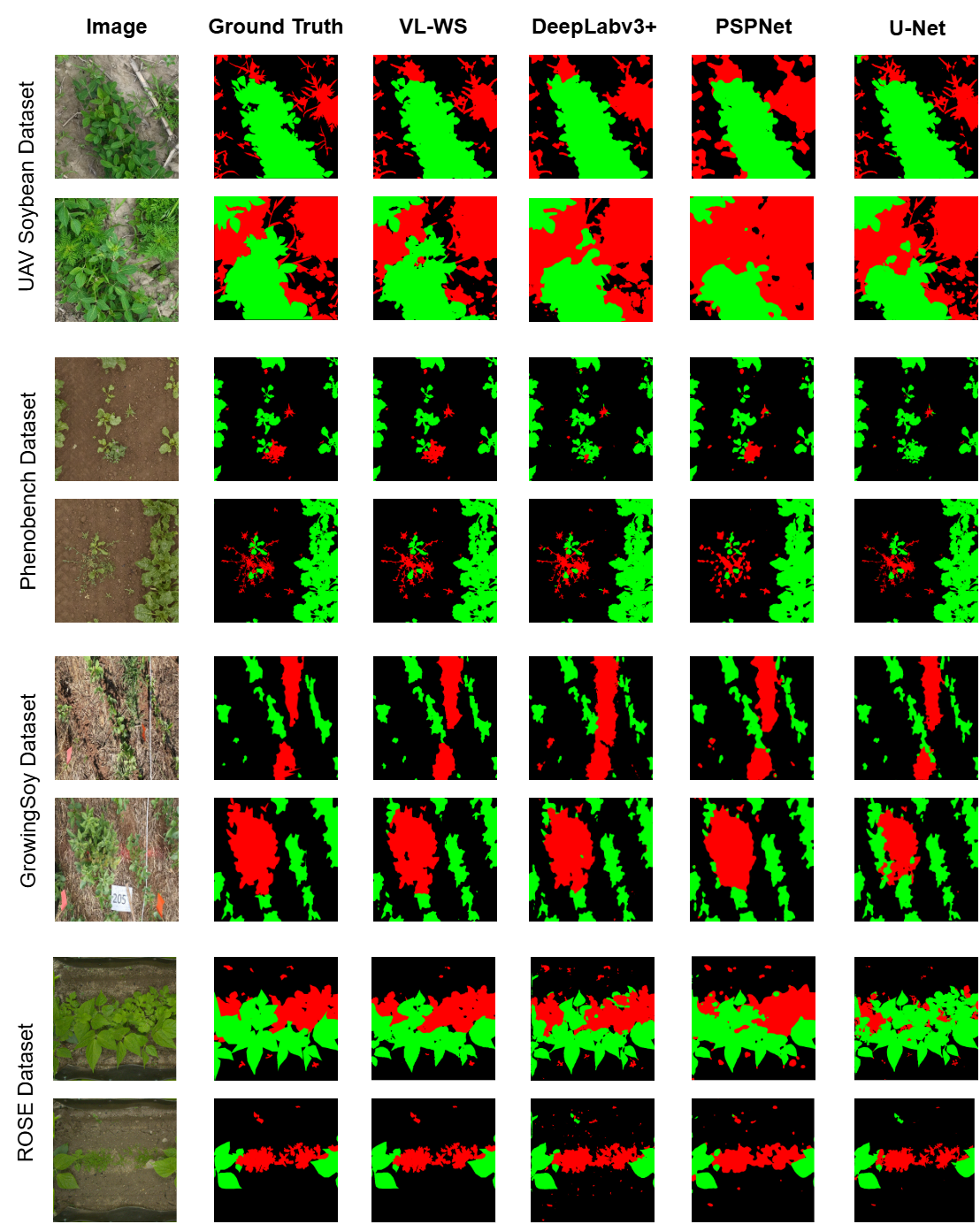}
    \caption{Qualitative comparison of segmentation results across datasets for the proposed model and baseline methods. Segmentation masks show crop in green, weed in red, and background in black. }
    \label{fig:comparison_model.drawio}
\end{figure*}

\begin{figure*}[t]
    \centering
    \includegraphics[width=0.8\textwidth]{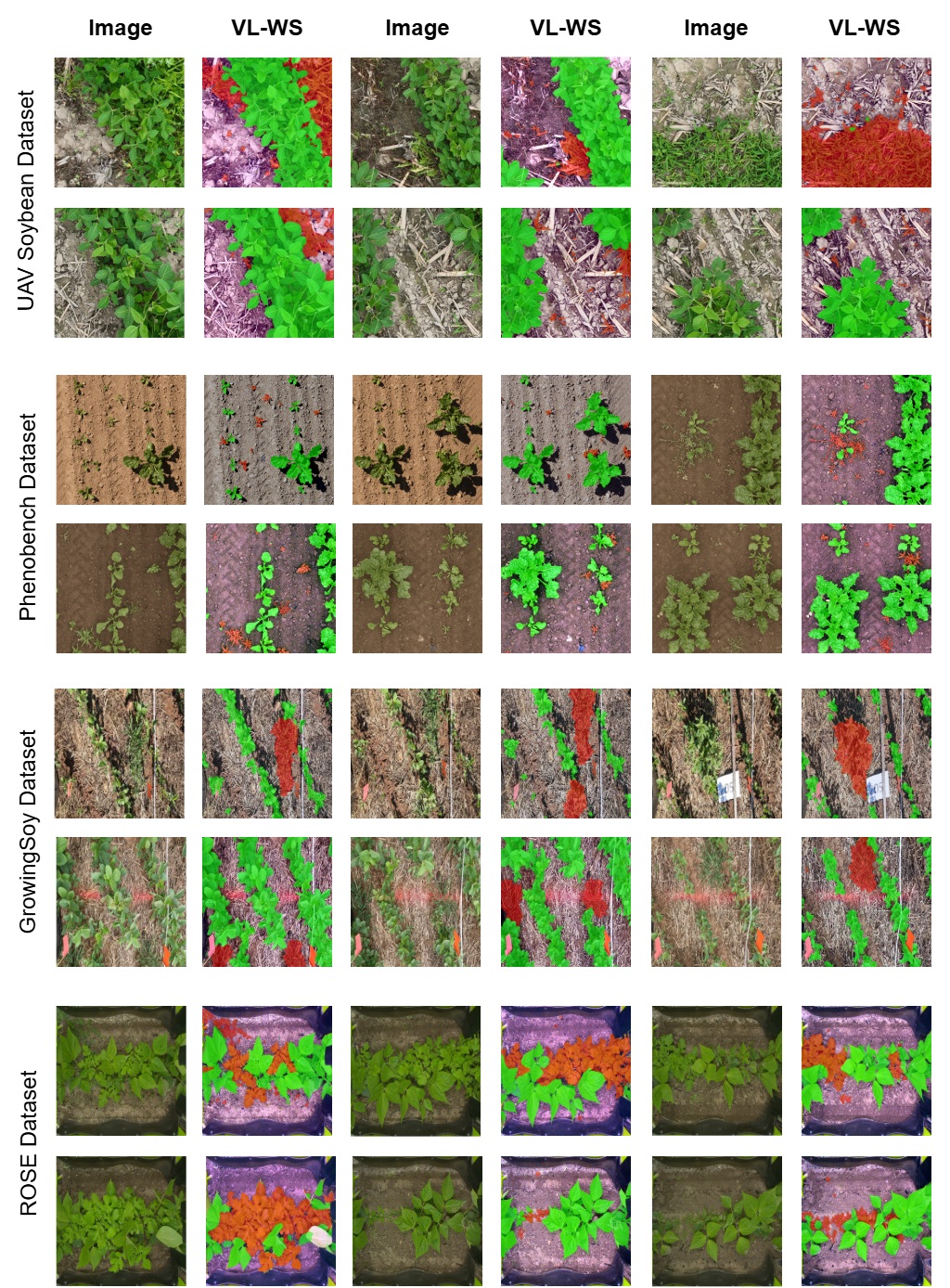}
    \caption{Qualitative segmentation results of the proposed VL-WS model across four agricultural datasets, demonstrating robust weed detection under varying lighting conditions, background textures, crop growth stages, and sensing platforms, including UAV and ground-based imagery. Predictions are overlaid on input images with green indicating crop, red indicating weed, and semi-transparent regions showing background.}
    \label{fig:predicted_images.drawio}
\end{figure*}

\begin{table}[h]
\centering
\caption{Ablation study on the vision-language contrastive loss weight ($\lambda_{\mathrm{VL}}$). Class-wise and mean Dice (\%) scores are reported across the multi-dataset benchmark.
}

\label{tab:loss_weight_singlecol}
\begingroup
\setlength{\tabcolsep}{3pt}
\begin{tabular}{lcccc}
\toprule
\textbf{$\lambda_{\text{VL}}$} &
\textbf{Weed} &
\textbf{Crop} &
\textbf{Bg} &
\textbf{Average} \\
\midrule
0.01 & 79.93 & 95.11 & 99.22 & 91.42 \\
0.02 & 80.45 & 95.23 & 99.24 & 91.64 \\
0.03 & 80.00 & 95.23 & 99.25 & 91.49 \\
0.05 & 80.41 & 95.33 & 99.25 & 91.66 \\
0.10 & 80.16 & 95.26 & 99.25 & 91.55 \\
\bottomrule
\end{tabular}
\endgroup
\end{table}

\subsection{Ablation Study on Vision–Language Loss}
Table \ref{tab:loss_weight_singlecol} reports segmentation performance under different values of the vision-language contrastive loss weight \(\lambda_{\mathrm{VL}}\). Weed Dice improves as \(\lambda_{\mathrm{VL}}\) increases from 0.01 to 0.02, reaching its peak at \(\lambda_{\mathrm{VL}} = 0.02\), after which performance plateaus or slightly declines. Crop segmentation exhibits marginal but consistent improvements with increasing \(\lambda_{\mathrm{VL}}\), while background Dice remains stable across all settings.

\subsection{Visual Analysis of Results}

\begin{figure*}[t]
    \centering
    \includegraphics[width=0.8\textwidth]{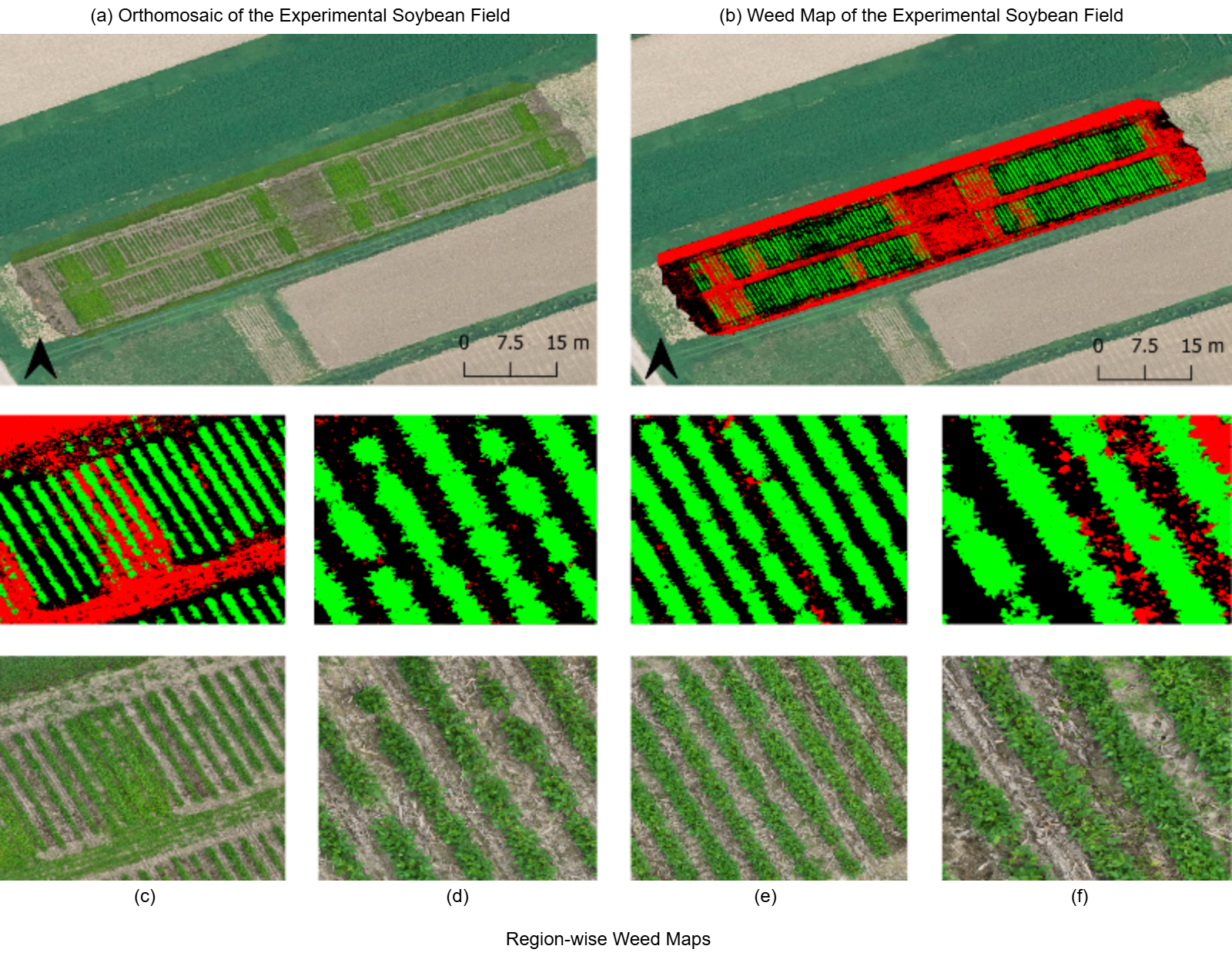}
    \caption{Field-scale weed segmentation results and zoomed predictions. (a) RGB orthomosaic of the soybean field. (b) Predicted weed distribution map with green indicating crop, red indicating weed, and black indicating background. (c)–-(f) Zoomed predictions from four field regions (top row) with corresponding RGB imagery (bottom row).}
    \label{fig:weedmap}
\end{figure*}

Fig.\ref{fig:comparison_model.drawio} presents qualitative comparisons with two representative sample images from each dataset, showing ground truth annotations alongside predictions from baseline methods and our model. Our model produces predictions that closely align with ground truth in spatial extent and boundary delineation. In densely interwoven crop-weed regions, baselines generate fragmented predictions with blurred boundaries and class leakage, while our approach produces sharper, more coherent segmentation. While baselines struggle with visually similar early-stage crops and weeds, our model effectively distinguishes between classes and maintains consistent performance across diverse crop types, weed species, and imaging conditions. Additional results in Fig. \ref{fig:predicted_images.drawio} illustrate robustness to data heterogeneity through prediction overlays. Examples span varied lighting conditions, background types, growth stages, and platforms (UAV and ground-based), with stable performance despite significant variation in GSD. Fig.\ref{fig:weedmap} shows field-scale weed maps generated by stitching predictions from individual image tiles. These maps reveal spatial weed distribution patterns across the entire field, identifying heavily infested regions and clean areas. Such spatially continuous representations enable targeted weed management for precision agriculture applications.

\begin{figure}
    \centering
    \includegraphics[width=\columnwidth]{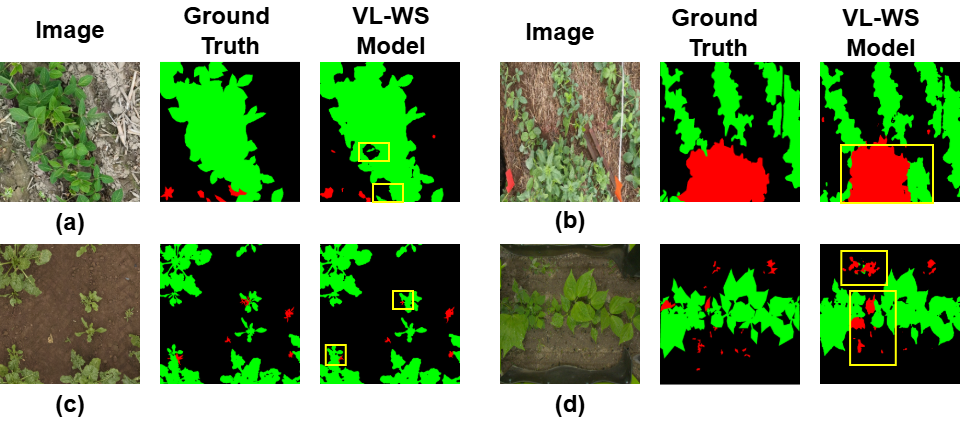}
    \caption{Qualitative error analysis of crop-weed segmentation predictions. Predicted segmentation masks are shown, with yellow boxes highlighting regions of inconsistency between model predictions and the ground truth, illustrating common failure cases under challenging field conditions.
}
    \label{fig:error_analysis}
\end{figure}

\paragraph{Error Analysis.}
Fig.~\ref{fig:error_analysis} presents a qualitative error analysis across four challenging cases. For each example, the input image, ground-truth annotation, and VL-WS model prediction are shown, with yellow boxes highlighting regions of discrepancy between prediction and ground truth. Examples (a) and (d) reveal annotation inconsistencies in the ground-truth labels. In (a), one of the highlighted regions is labeled as crop in the ground truth but predicted as background by the model. Visual inspection of the input image indicates that this region corresponds to bare soil visible between leaf gaps, suggesting that the model prediction is more consistent with the actual image. Similarly, in (d), the model predicts weed pixels in a region labeled as background in the ground truth, while the input image shows clear weed presence, indicating a potential labeling error. These cases highlight a common challenge in crop-weed datasets. Fine-scale plant morphology, overlapping vegetation, and complex leaf structures make precise pixel-level annotation difficult and can introduce label noise. Examples (a), (b), and (c) illustrate actual error cases driven by visual ambiguity. In these regions, crops and weeds exhibit highly similar appearance with overlapping foliage and poorly defined boundaries. As a result, the model struggles in ambiguous transition zones where discriminative visual cues are limited.

\subsection{Discussion}
Experimental results demonstrate that, in the multi-dataset setting, the proposed VL-WS model consistently outperforms standard CNN-based architectures for weed segmentation, highlighting the effectiveness of vision-language semantic grounding over purely visual feature learning in agricultural segmentation tasks. Several factors contribute to this advantage, most notably the use of a frozen CLIP encoder, which stabilizes optimization by preventing the representation space from drifting as heterogeneous gradients are applied. This design decouples semantic understanding from spatial localization, allowing the model to rely on pretrained vision-language semantics for identifying crop and weed concepts while learning precise boundaries through task-specific layers. As a result, the model reduces its dependence on costly pixel-level annotations and improves data efficiency. Additionally, FiLM-based modulation enables context-aware adaptation by selectively enhancing semantically relevant features while preserving the shared semantic knowledge from vision-language pretraining, allowing the model to accommodate dataset-specific variation without compromising cross-domain transferability.

Beyond comparisons with standard CNN baselines in the multi-dataset setting, the proposed framework further demonstrates competitive performance when compared to prior state-of-the-art methods evaluated on single-dataset agricultural benchmarks. In the PhenoBench dataset, which provides standardized benchmarks for semantic segmentation alongside other plant perception tasks \citep{10572312}, existing results are reported for DeepLabV3+ and ERFNet. DeepLabV3+ achieves IoU scores of 64.59\% (weed), 94.07\% (crop), and 99.25\% (soil), while a Bayesian DeepLabV3 variant with Monte Carlo dropout reports IoU values of 63.37\% (weed), 94.60\% (crop), and 99.33\% (background) on the validation set \citep{celikkan2023semantic}. In comparison, VL-WS achieves Dice scores of 77.57\% (weed), 95.84\% (crop), and 99.49\% (background). After conversion from Dice to IoU, these correspond to 63.35\% (weed), 92.01\% (crop), and 98.98\% (background), demonstrating performance comparable to established state-of-the-art methods. Importantly, this performance is achieved under a multi-dataset training setting, whereas the reported benchmarks are obtained under single-dataset training.

A comparable trend is observed on the ROSE dataset, which is commonly used to evaluate robustness under environmental variability and domain shift. Prior work by \citet{catalano2024tackling} reports a weed IoU of 60.39\% in bean fields when trained on the full ROSE dataset. In our experiments, VL-WS is evaluated on a bean subset containing four weed species, split into training, validation, and test sets. On the validation set, VL-WS achieves a weed Dice score of 75.66\%, corresponding to an IoU of 60.84\%, which is closely aligned with the reported benchmark. For the GrowingSoy dataset, most prior studies focus on instance segmentation using YOLOv5 and YOLOv8 variants \citep{steinmetz2024seedling}, limiting direct comparison with semantic segmentation approaches. Nevertheless, the strong performance of VL-WS on GrowingSoy further supports its robustness across datasets with differing acquisition conditions. Collectively, these comparisons show that the proposed language-guided approach maintains competitive performance across both single and multi-domain benchmarks, matching single-dataset performance even under joint training on heterogeneous datasets, unlike purely visual models that suffer from negative transfer.

\begin{figure}
    \centering
    \includegraphics[width=\columnwidth]{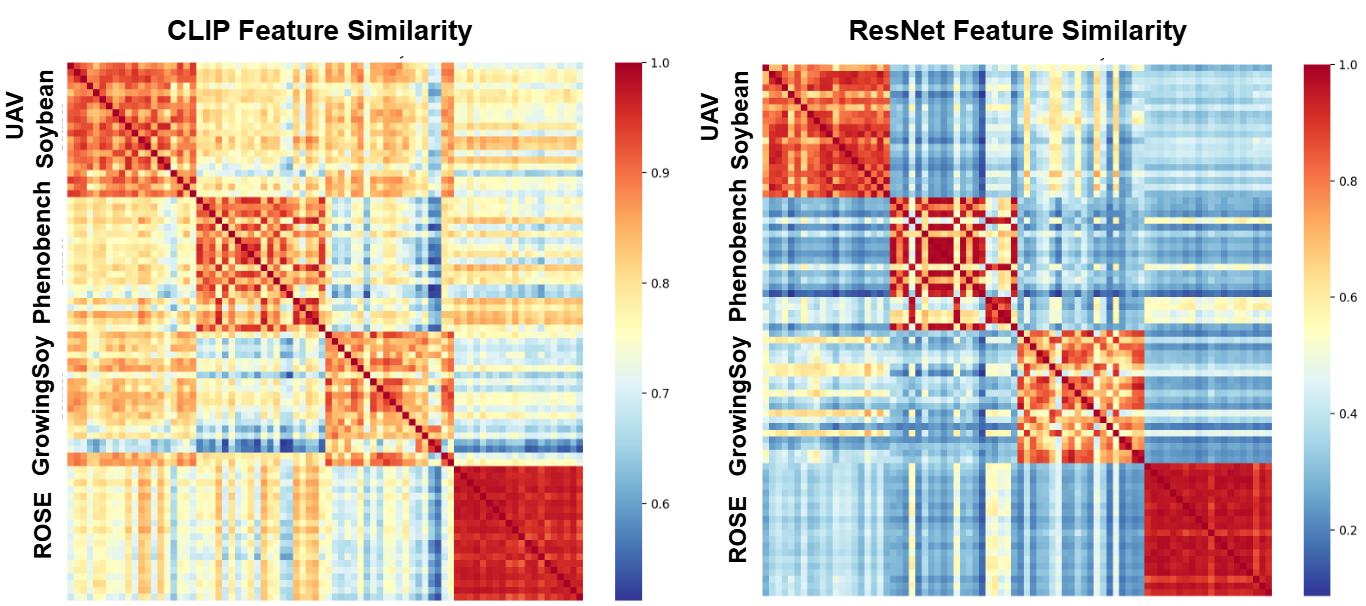}
    \caption{Cross-dataset cosine similarity of deep feature embeddings from pretrained CLIP and ResNet encoders. Heatmaps showing pairwise cosine similarity between feature embeddings of image tiles drawn from four weed segmentation datasets: UAV Soybean, Phenobench, GrowingSoy, and ROSE. (Left) CLIP encoder features exhibit high within-dataset and substantial cross-dataset similarity, reflecting semantically rich and more dataset-invariant representations. (Right) Features extracted from a ResNet encoder (pretrained on ImageNet) display strong clustering within datasets but markedly lower similarity across datasets, indicating limited generalizability. The cross-dataset feature alignment achieved by the CLIP encoder supports improved robustness in multi-domain weed segmentation tasks.
}
    \label{fig:cosine_similarity.drawio}
\end{figure}

As shown in Fig. \ref{fig:cosine_similarity.drawio}, CLIP features exhibit strong similarity both within and across datasets, indicating a semantically consistent representation space. In contrast, features from an ImageNet-pretrained ResNet show strong within-dataset similarity but substantially lower cross-dataset similarity, reflecting sensitivity to dataset-specific visual characteristics. This contrast suggests that CLIP representations encode higher-level semantics that generalize across crops, weed species, and imaging conditions, whereas purely visual features remain more tightly coupled to domain-specific appearance. This behavior is consistent with prior studies showing that image-only encoders tend to rely on low-level cues such as texture and color, while language supervision encourages representations organized around semantic concepts that are more robust to appearance variation.

Overall, these results, together with quantitative comparisons to prior work and the embedding space analysis, demonstrate that incorporating linguistic priors encourages a more semantically structured organization of the representation space, reducing reliance on low-level visual cues. By shifting toward semantically grounded representations, the VL-WS framework supports more domain-agnostic segmentation behavior and mitigates sensitivity to dataset-specific appearance variations. This design reduces dependence on exhaustive, site-specific pixel-level annotations and facilitates more effective learning under heterogeneous agricultural conditions.

Overall, these quantitative and qualitative analyzes demonstrate that linguistic priors induce a more structured representation space, reducing the reliance on low-level visual cues. By integrating semantically grounded features, VL-WS achieves domain-agnostic segmentation and reduced sensitivity to dataset-specific appearance shifts. Consequently, this design minimizes reliance on exhaustive, site-specific annotations and enables robust learning across heterogeneous agricultural environments.

\section{Conclusion}
We demonstrate that vision-language semantic grounding effectively addresses negative transfer and label heterogeneity in multi-dataset agricultural segmentation. By combining frozen CLIP representations with a learnable spatial encoder and caption-conditioned feature modulation, VL-WS decouples semantic understanding from spatial localization and substantially outperforms standard CNN baselines in weed segmentation, while maintaining strong performance under limited supervision. These results highlight vision-language alignment as a promising foundation for robust and generalizable segmentation across diverse agricultural environments. While the proposed architecture reduces performance degradation arising from intra-class variation across heterogeneous datasets, it does not fully eliminate negative transfer, as the trainable spatial encoder and decoder remain susceptible to conflicting visual patterns from morphologically diverse weed species and varying imaging conditions during joint training. In addition, reliance on global image-level vision-language embeddings may limit the capture of fine-grained, spatially localized semantic distinctions in densely interwoven crop-weed regions. Building on the proposed framework, future work could further mitigate residual negative transfer by introducing stronger semantic regularization within the spatial encoder, as well as spatially adaptive vision-language conditioning to better align semantic cues with visual features. The framework could also be extended to temporal learning settings, where multi-stage growth supervision captures morphological divergence between crops and weeds. Incorporating temporal cues together with multispectral or multi-modal inputs may further improve robustness under phenological variation and challenging field conditions.

\section*{CRediT authorship contribution statement}
\textbf{Nazia Hossain:} Data curation, Investigation, Methodology, Visualization, Formal analysis, Writing - original draft,  Writing - review \& editing.
\textbf{Xintong Jiang:} Data curation, Investigation, Writing - review \& editing.
\textbf{Yu Tian:} Writing – review \& editing.
\textbf{Philippe Seguin:} Resources, Writing - review \& editing.
\textbf{O. Grant Clark:} Writing - review \& editing.
\textbf{Shangpeng Sun:} Conceptualization, Resources, Writing - review \& editing, Supervision, Project administration, Funding acquisition.

\section*{Declaration of competing interest}
The authors declare that they have no known competing financial interests or personal relationships that could have appeared to influence the work reported in this paper.
\section*{Acknowledgments}
This research is supported by funding from the FRQNT \& MAPAQ Partnership Research Program-Sustainable Agriculture (Grant No. 259806), the McGill Collaborative for AI and Society Interdisciplinary Research Program (Grant No. 173165), and the RQRAD Emerging Project (Grant No. 265224). We thank Dr.Huong Nguyen for critically proofreading the manuscript.

\section*{Data availability}
Data will be made available on request.
\FloatBarrier

\bibliographystyle{cas-model2-names}

\bibliography{cas-refs}
\end{document}